\pdfoutput=1

\documentclass[11pt]{article}

\usepackage{acl}

\usepackage{times}
\usepackage{latexsym}

\usepackage[T1]{fontenc}

\usepackage[utf8]{inputenc}

\usepackage{microtype}

\usepackage{amsfonts}
\usepackage{graphicx}
\usepackage{booktabs}
\usepackage{hyperref}
\usepackage{multirow}

\title{Relational Graph Convolutional Neural Networks for Multihop Reasoning: A Comparative Study}

\author{Ieva Stali\={u}nait\.{e}\thanks{$\; $ Now at Department of Computer Science and Technology, University of Cambridge}, Philip John Gorinski, Ignacio Iacobacci \\
    Huawei Noah’s Ark Lab, London \\ 
       \texttt{irs38@cam.ac.uk} \\
       \texttt{\{philip.john.gorinski,ignacio.iacobacci\}@huawei.com}
    }

\begin{document}
\maketitle
\begin{abstract}
Multihop Question Answering is a complex Natural Language Processing task that requires multiple steps of reasoning to find the correct answer to a given question. Previous research has explored the use of models based on Graph Neural Networks for tackling this task. Various architectures have been proposed, including Relational Graph Convolutional Networks (RGCN). For these many node types and relations between them have been introduced, such as simple entity co-occurrences, modelling coreferences, or ``reasoning paths'' from questions to answers via intermediary entities. Nevertheless, a thoughtful analysis on which relations, node types, embeddings and architecture are the most beneficial for this task is still missing. In this paper we explore a number of RGCN-based Multihop QA models, graph relations, and node embeddings, and empirically explore the influence of each on Multihop QA performance on the WikiHop dataset.
\end{abstract}

\section{Introduction}

As the latest models on question answering tasks have high performance on most popular benchmarks, datasets for the more complex task of multihop question answering have been introduced \cite{welbl2018constructing, yang2018hotpotqa}. The multihop question answering task requires a model to provide an answer given a question and multiple background documents. The questions require reasoning across more than one document for finding the correct answer. For example, to answer the question given in Figure~\ref{fig:question} one needs to reason over background documents 1 and~3.

\begin{figure}
    \textbf{Query}: \textit{applies\_to\_jurisdiction vice president of the European Parliament}
    
    \vspace{.25em}
    \textbf{Background document 1}: \textit{The European Parliament (EP) is the directly elected parliamentary institution of the European Union (EU) [...]}
    
    \vspace{.25em}
    \textbf{Background document 2}: \textit{The European Commission (EC) is an institution of the European Union, responsible for proposing legislation [...]}
    
    \vspace{.25em}
    \textbf{Background document 3}: \textit{There are fourteen vice-presidents of the European Parliament who sit in for the president in presiding over the plenary of the European Parliament [...]}
    
    \vspace{.25em}
    \textbf{Answer}: \textit{European Union}
    \caption{Example WikiHop question requiring reasoning over several documents to find the correct answer.}
    \label{fig:question}
\end{figure}

In order to solve this problem in an informed way the system needs to recognize entities, resolve their coreference, understand the relations between them, and choose the right entity as the answer, or answer the question in free form. 

People perform this task without too much difficulty, for example when playing the \textit{Wikiracing}\footnote{\url{en.wikipedia.org/wiki/Wikiracing}} game.
We posit that what a human does is start with a given entity (e.g. \textit{Vice President of the European Parliament}), traverse Wikipedia articles by choosing entities in the topic of the question (e.g. \textit{European Parliament}) and finally find the answer entity in that article by searching for the entity that links to the original entity in the question through the relation mentioned. 

Transformer-based language models fine-tuned for QA tasks, including Multihop QA, reach very high performance on many benchmarks without explicitly modeling the multiple reasoning hops~\cite{beltagy2020longformer, he2020realformer}. However, they are very computationally expensive as they model relations between all tokens in the input with attention. This motivates the study of which specialized relations would be the most efficient at finding the multihop reasoning paths.

To this end, a variety of Graph Neural Network (GNN) approaches have been proposed in recent years, which explore the use of graph structures with entities as vertices and relations as edges to address the Multihop QA task. These model the relationship between the question, context, and potential answers in a more informed way than full token attention.

In particular, Relational Graph Convolutional Networks \cite{schlichtkrull2018modeling} (RGCN) have been successfully applied in a number of Multihop QA models. RGCN introduces typed edges between nodes in the underlying graph, with a convolutional update step of nodes depending on their neighbours and their respective type(s) of relation. 

While the general architecture of RGCN is shared by many approaches, the remaining framework (node types, embeddings, relations, pre- and post-RGCN layers) vary greatly. However, a principled analysis of the impact these factors have on model performance seems still to be missing from current literature. This paper aims at rectifying this by shedding some light on the efficacy of RGCN for multihop reasoning under various conditions.

The main contributions of this paper are as follows: (i) We present a direct comparison of two strong, related, yet significantly different RGCN-based architectures for the Multihop QA task; (ii) we derive a new RGCN-based architecture combining features of the two prior ones; (iii) we present a principled analysis of the impact on model performance across three conditions: model architecture, node types and relations, node embeddings.

To our knowledge, this is the first attempt at a principled comparison of RGCN-based Multihop QA approaches.


\section{Related Work}

A number of graph-based approaches to Multihop QA have been proposed in recent years.
To answer the questions presented in the WikiHop dataset \cite{welbl2018constructing}, in \citet{de2019question} the authors propose to use Relational Graph Convolutional Networks \cite{schlichtkrull2018modeling} by modelling relations between entities in the query, documents, and candidate choices. Path-based GCN was introduced by \citet{tangmulti} for the same task and builds on these relations, but added relations over ``reasoning entites'', i.e., Named Entities that co-occur with those presented in the query and candidate entity set.
Furthermore, \citet{tu-etal-2019-multi} have used different types of nodes and edges in the graph which led to high performance. They employed entity, sentence and document level nodes to represent the relevant background information and connect them on the basis of co-reference in the case of entities and co-occurence among all types of nodes. 
In a similar vein, in order to answer HotpotQA \cite{yang2018hotpotqa} questions, which include answer span as well as support sentence prediction tasks, \citet{fang-etal-2020-hierarchical} introduced Hierarchical Graph Networks, which establish a relational hierarchy between graph nodes on entity, sentence, and paragraph levels, and uses Graph Attention Networks \cite{velivckovic2017graph} for information distribution through the graph.
HopRetriever \cite{DBLP:journals/corr/abs-2012-15534} leverage Wikipedia hyperlinks to model hops between articles via entities and their implicit relations introduced through the link.

In contrast, some research has explored the question of whether a multihop approach is even necessary to solve the task presented in recent multihop question answering datasets \cite{min2019compositional}. They show that using a single hop is sufficient to answer 67\% of the questions in HotpotQA. What is more, \citet{tang2020multi} study the ability of state-of-the-art models in the task of Multihop QA to answer subquestions that compose the main question. They show that these models often fail to answer the intermediate steps, and suggest that they may not actually be performing the task in a compositional manner. Furthermore, \citet{groeneveld2020simple} provide support for the claim that the multihop tasks can be solved to a large extent without an explicit encoding of the compositionality of the question and all the relations between knowledge sources in different support documents in HotpotQA. Their pipeline simply predicts the support sentences separately and predicts the answer span from them using transformer models for encoding all inputs for the classification. 
In a similar vein, \citet{shao2020graph} show that self-attention in transformers performs on par with graph structure on the HotpotQA task, providing further evidence that this dataset does not require explicit modeling of multiple hops for high performance. 

\section{QA Graphs and Architecture}
\paragraph{WikiHop}
A number of Multihop QA datasets have been released with the largest two, WikiHop \citep{welbl2018constructing} and HotpotQA \citep{yang2018hotpotqa}, most actively used for research. The precise nature of the task differs greatly between datasets -- Wikihop requires selecting the right answer entity from a list of candidates; HotpotQA requires a multi-task setup of answer span prediction (without candidates) and support sentence prediction -- and they require specialised architectures to address them. This comparative study can for lack of space only focus in-depth on model performance on one dataset. We chose the WikiHop, as a variety of different-yet-related Graph Neural Network architectures have been proposed for it.

In WikiHop, instances consist of a \emph{query} -- represented as a tuple $q=<\!\!s,r,?\!\!>$ of subject entity $s$, unknown answer entity $?$, and a relation $r$ that holds between them -- a set of candidate entities (one of which is the correct answer to the query), and a number of background documents that may or may not hold relevant information to identify the correct candidate entity as the answer. As illustrated in Figure~\ref{fig:question}, multiple ``hops'' through the background documents are usually required to identify the correct answer, i.e., WikiHop presents a complex reasoning task in which multiple sources of information need to be used and related to achieve the goal. This makes it an ideal dateset for analysing the multihop reasoning capabilities of various model architectures under varying available information.

\paragraph{Relational Graphs and RGCN}
A relational graph $G=(V,E^R)$ consists of a set of vertices $V$ and a set $E^R$ of relational edges $(v_i, v_j, r)$, indicating specific relationships that apply between node in the graph. For Multihop Question Answering graphs, let vertices be the union $V=M_q\cup M_c\cup M_r$ of sets representing respectively all mentions that occur in the background documents belonging to (1) entities appearing in the query $q$; (2) a set of potential answer candidates; (3) the set of all Named Entities other than query and candidates (reasoning entities), and let $Q$, $C$, $E$ be the sets of unique referents of question, candidate, and reasoning entities. Furthermore, let $S$ be the set of all sentences in all documents, and $D$ the document set where each element of $D$ contains those sentences in $S$ which belong to the same background document. Let $|m|$ denote the string of a mention $m$, $||m||$ the unique referent of a mention, and $s(m),\ d(m)$ the sentence and document $m$ occurs in respectively.

Relational Graph Convolutional Networks were first introduced in \citet{schlichtkrull2018modeling} as an extension to Graph Convolutional Networks \cite{duvenaud2015convolutional, kipf2016semi}. Intuitively, (R)GCN can be seen as message passing systems \cite{gilmer2017neural} where information ``flows'' from nodes in a graph to their direct neighbours, possibly over multiple layers (or hops). At each layer, the current information about a node is updated by all its neighbours, and this update happens simultaneous for all nodes. Formally, given a relational graph $G=(V,E^R)$, we can define the Relational Graph Convolution neighbourhood $N_i^r$ of a vertex $v_i\in V$ for relation $r$ as $N_i^r=\{v_j | (v_j, v_i, r)\in E\}$, that is, it contains all incoming connections to $v_i$ under relation $r$. When vertices are represented as vectors $h\in \mathbb{R}^d$, e.g., via some node embedding function, we obtain the update $u_i^l$ for node $h_i^l$ at layer $l$ of the RGCN via the messages $m_i^l$ it receives from its neighbours:
\begin{equation}
    m_i^l = \sum_{r\in R}\sum_{j\in N_j^r}\frac{1}{|N_i^r|}W_r^lh_j
\end{equation}
\begin{equation}
    u_i^l = W_u^lh_i^l+m_i^l
\end{equation}
To arrive at the representation $h_i^l+1$, previous work has established the use of a gating mechanism \cite{tu-etal-2019-multi,de2019question,cao-etal-2019-bag}:
\begin{equation}
    a_i^l = \sigma (W_a[u_i^l;h_i^l]+b_a)
\end{equation}
\begin{equation}
    h_i^{l+1} =  a_i^l \odot \tanh(u_i^l) + (1-a_i^l) \odot h_i^l
\end{equation}

In addition to gating the node representation, query-aware gating \cite{tangmulti} allows to use query information in each node update. For query-aware gating, first the query $q$ of length $m$ is embedded into $p\in \mathbb{R}^{m\times d}$, for example with a bi-LSTM. Then a representation $q_i^l$ is obtained and used to inform the node update $u_i^l$ as
\begin{equation}
    q_{ij} = \sigma(W_q[u_i^l;p_j]+b_q)
\end{equation}
\begin{equation}
    \alpha_{ij} = \frac{\exp (q_{ij})}{\sum_{k=1}^{m}\exp(q_{ik})}
\end{equation}
\begin{equation}
    q_i^l=\sum_{j=1}^m \alpha_{ij}p_j
\end{equation}
\begin{equation}
    \beta_i^l=\sigma(W_\beta [q_i^l;u_i^l] + b_\beta)
\end{equation}
\begin{equation}
    u_i^l = \beta_i^l\odot \tanh(q_i^l) + (1-\beta_i^l)\odot u_i^l
\end{equation}
where $W_r^l, W_u^l \in \mathbb{R}^{d\times d}$, $W_a, W_q, W_\beta\in \mathbb{R}^{2d\times d}$.

\begin{figure*}[t]
    \centering
    \includegraphics[width=.85\textwidth]{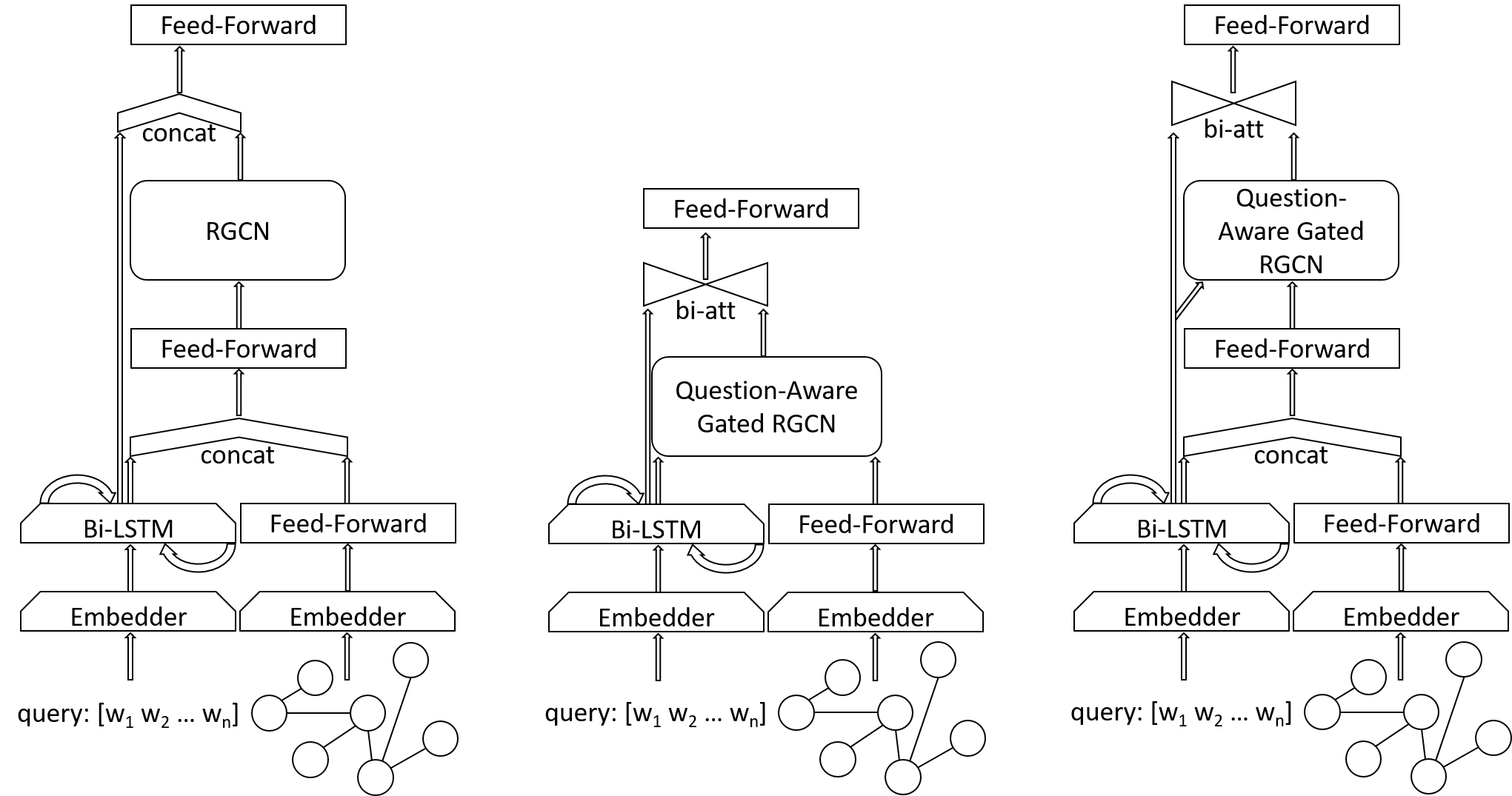}
    \caption{Outline Diagrams for EntityGCN (left), PathGCN (center), and MashupGCN (right).}
    \label{fig:entitypathgcn}
\end{figure*}

\paragraph{Model Architectures}
Many architectures have been proposed for the task. As the full body of proposed architectures would go beyond the constraints of this paper, we choose to focus on one well-established two RGCN-based models, \emph{EntityGCN} \cite{de2019question} and \emph{PathGCN} \cite{tangmulti}, in this study. Diagrams of the models are given in Figure~\ref{fig:entitypathgcn} left (EntityGCN) and center (PathGCN). For space constraints, we refer readers to the original papers for exact model details.

Both EntityGCN and PathGCN follow a similar processing pipeline: First, a graph is generated based on a WikiHop instance. Named Entity Recognition (NER) is performed on the question and documents, and entities are extracted as nodes and connected with typed edges. Query and nodes are embedded through an embedding function, and the query embeddings are further processed with a BiLSTM \cite{hochreiter1997long}.
\\
After this step, further processing differs: EntityGCN concatenates the query representation to all node representations, and generates a joint embedding with another feed-forward layer. The resulting graph is used in a multi-layer RGCN to update the nodes, which are then concatenated with the query again to obtain the final entity node representations.
\\
PathGCN uses query and initial node representations as input to a question-aware gated RGCN, which updates the nodes based on their neighbours with the help of question-aware and output gates. The resulting updated node representations along with the query are passed through a bi-directional attention layer \cite{cao-etal-2019-bag} to obtain the final entity node representations.
\\
Finally, both EntityGCN and PathGCN employ a softmax output layer over the entity graph, to determine the final answer entity.

In addition to EntityGCN and PathGCN, we also introduce a novel \emph{MashupGCN} for our analysis, shown in Figure~\ref{fig:entitypathgcn} (right). We combine features of the two models: Before the RGCN, we concatenate query and nodes like \citet{de2019question}, but then employ a question-aware RGCN and bi-directional attention like \citet{tangmulti}.

\paragraph{Graph Nodes and Edge Types}
In addition to their architectures, \citet{de2019question} and \citet{tangmulti} further differ in the graph construction phase, with the latter building upon the former. For EntityGCN, the authors find initial nodes for a WikiHop example existing of query $q=<\!\!s,r,?\!\!>$, a set of answer candidates $C_q$, and a set of supporting documents $S_q$ by finding exact matches of $C_q\cup \{s\}$ in $S_q$. After this, they employ a coreference resolution system to discover further mentions of candidates and question entities in the support documents. Finally, they define three relations over the discovered nodes: (i) Nodes that co-occur in the same document; (ii) exact-match edges for matching named entities; (iii) coreference edges for nodes that are part of the same coreference chain.

For PathGCN, the authors expand on this construction. In addition to entities in $C_q\cup \{s\}$, they further add \emph{reasoning entities} to the graph. Reasoning entities are those that can be collected along a ``reasoning path'' from a query entity to an answer candidate entity through the supporting documents. Accordingly, the authos define three additional relational edges: (iv) query-reason to connect query nodes to reasoning entity nodes from the same sentence in the same document; (v) reason-reason for two adjacent reasoning nodes on the same path; (vi) reason-cand, connecting reasoning and candidate nodes from the same document sentence.
\\
In both cases, the authors also define a final additional edge type (vii) complement, connecting all nodes that are in no other relation.

It is worth noting that for PathGCN, nodes of type (iii) are omitted as no coreference resolution is performed. Since even for EntityGCN, using coreferent nodes and edges is shown to be of low impact while adding significant processing overhead for their resolution, we adapt this latter approach and also drop the notion of proper coreference. Instead, we re-interpret relations (ii) as exact matches \emph{across} documents -- that is, connected nodes stem from different support documents -- and relation (iii) as exact matches \emph{within} documents, i.e., ``coreference'' chains of string-matching entities.

\paragraph{Question and Node Embeddings}
As mentioned before, after the graph construction phase the query and graph nodes are embedded via some embedding function. \citet{de2019question} use ELMO \citep{peters2018deep} to embed the nodes as well as the query before the BiLSTM application, while \citet{tangmulti} embed the query with GloVe \cite{pennington2014GloVe} before the BiLSTM, while graph nodes are embedded with GloVe and ELMO, and their representations concatenated.

We make the embedding type and its impact on multihop reasoning capabalities a subject of our analysis. In order to better assess it, we will follow EntityGCN's approach of using the same embedder for query and nodes, rather than always sticking to GloVe for query representations. In addition to the mentioned types, we also consider transformer-based \cite{vaswani-etal-2017-attention} representations in the form of RoBERTa embeddings \cite{liu2019roberta}.

\section{Experiments}

We carried out a battery of experiments to test how well RGCN-based models learns to answer multihop questions under various conditions. To this end, we re-implemented the models of \citet{de2019question} and \citet{tangmulti}, as well as our own MashupGCN, as introduced in the previous section. We trained and evaluated each model on the WikiHop training and development sets respectively, while altering the kind information available to them. EntityGCN and PathGCN were trained with the hyperparameters presented in their respective papers, for MashupGCN we used the same hyperparameters as EntityGCN.

\paragraph{Impact of Model Architecture and RGCN}
First, we are interested in the effect that the different architectures have on performance, as well as the impact the RGCN has in the first place. Table~\ref{tab:res_arch_rgcn} summarizes the results of these experiments.

\begin{table*}[t]
    \small
    \centering
    \begin{tabular}{l|c|c|c|c|c|c}
        \toprule
         & EntityGCN & EntNoGCN & PathGCN & PathNoGCN & MashupGCN & MashupNoGCN\\
         \midrule
         Accuracy & 65.77 & 62.52 & 64.63 & 64.30 & 67.01 & 62.34 \\
         \bottomrule
    \end{tabular}
    \caption{Performance of different Multihop QA models with and without RGCN, using query and candidate nodes with corresponding relations and ELMO embeddings.}
    \label{tab:res_arch_rgcn}
\end{table*}

For the results in Table~\ref{tab:res_arch_rgcn} we employed the embeddings and node/relation types of EntityGCN, as these are the ``smallest common denominator'' between all models, while the original PathGCN paper introduces more nodes and relations, and changes embeddings. We note that our results for EntityGCN are broadly consistent with the numbers in the original paper, which reported an accuracies if $65.1$ and $62.4$ with and without RGCN respectively. PathRGCN performs slightly worse in this experiment however, we point out that this setting differs significantly from that in the original paper, omitting reasoning entities and the corresponding relations.

While not surprising, it is worth noting that the RGCN is crucial for the question answering performance. All models show a substantial loss in answer prediction accuracy when the GCN is skipped.

\paragraph{Impact of Nodes and Relational Edges}
\begin{table*}[ht]
    \small
    \centering
    \begin{tabular}{l|c|c|c||c|c|c}
        \toprule
         & EntityGCN & PathGCN & MashupGCN & EntityGCN+ & PathGCN+ & MashupGCN+\\
         \midrule
         Base & 65.77 & 64.63 & 67.01 & -- & -- & -- \\
         +Reason & 67.77 & 66.67 & 67.10 & -- & -- & -- \\
         +Sents & 66.08 & 65.85 & 66.20 & -- & -- & -- \\
         +Reason+Sents & 66.93 & 66.54 & 64.85 & 67.83 & 66.71 & 67.28 \\
         \bottomrule
    \end{tabular}
    \caption{Answer prediction accuracy of model architectures given various node and edge types. Base refers to only query and candidate nodes, +Reason adds reasoning entities, +Sents adds sentence nodes and relations. The final row combines all nodes and edges. GCN+ denote models with scaled-up parameters.}
    \label{tab:res_graph}
\end{table*}

Next, we explore model performance under changing conditions of nodes and their relations used in the constructed graph. In Table~\ref{tab:res_graph}, we show results of this experiment.

As before, we use ELMO embeddings for these experiments to compare across models. In Table~\ref{tab:res_graph}, ``Base'' refers to a graph with only query and candidate entitites, as described in the original EntityGCN paper. The ``+Reason'' setting adds the reasoning entitites and relations introduced for PathGCN. We point out that in this condition, our numbers broadly reflect those reported in \citet{tangmulti} (66.1) when using ELMO embeddings.

For ``+Sents'', we introduce additional nodes into the graph that encode support document \emph{sentences}, and add according edges. These additional edges encode connect (viii) two sentence nodes that stem from the same document; (ix) two subsequent sentences; (x) a sentence to its immediate predecessor; (xi) a sentence to its immediate successor; (xii) sentences to the nodes of entities they contain. The final row in Table~\ref{tab:res_graph} denotes the full set of entity and sentence nodes with corresponding relations.

For EntityGCN and PathGCN, both the addition of reasoning entities and relations as well as adding sentences to the graph have a positive impact on performance, with an additional $~2$ and $~1$ point accuracy over the base setting respectively. Interestingly, MashupGCN does not seem to similarly benefit from the additions, performing virtually identically when reasoning entities are added, and even dropping by almost $1$ point with additional sentence nodes.

For EntityGCN and PathGCN models, we further observe that combining reasoning and sentence nodes in the graph improves over the base graph and sentences, but falls short of the performance of using a reasoning graph without sentence nodes. We hypothesize that the models may be under-parametrized for this setting: The number of nodes in a graph increases significantly when adding reasoning and sentence nodes: On average, the WikiHop training data leads to graphs of around 59 nodes in the base setting (query+candidate entities). This number almost doubles to 112 when adding sentence nodes. Finally, the average number of nodes when considering all entities (query, candidate, and reasoning) rises to 300.

To validate this intuition, we perform additional experiments with scaled-up versions of the three models, doubling the number of parameters. Results seem to be consistent with our assumption, as the larger models exhibit better performance when training with the full set of nodes and relations, as shown on the right-hand side of Table~\ref{tab:res_graph}.

\paragraph{Impact of Word Embedding Type}

\begin{table*}[t]
    \small
    \centering
    \begin{tabular}{l|c|c|c|c|c|c}
        \toprule
         & \multirow{2}{*}{GloVe} & \multirow{2}{*}{ELMO} & \multirow{2}{*}{RoBERTa} & \multirow{2}{*}{GloVe+ELMO} & \multirow{2}{*}{ELMO+RoBERTa} & GloVe+ELMO\\
         & & & & & & +RoBERTa \\
        \midrule
        EntityGCN & 61.93/65.05 & 65.77/67.77 & 63.83/64.18 & 65.63/68.03 & 65.34/67.87 & 66.28/68.01 \\
        PathGCN & 61.44/63.20 & 64.63/66.67 & 61.48/63.10 & 64.38/66.52 & 64.61/65.26 & 64.83/67.05\\
        MashupGCN & 61.95/64.57 & 64.97/67.10 & 63.46/64.91 & 66.56/67.73 & 66.01/68.05 & 67.38/67.87 \\
        \bottomrule
    \end{tabular}
    \caption{Accuracy of models trained with different embeddings and their combinations, tested on the development set of WikiHop. Models using query and candidate entities (left number), and additional reasoning entities (right).}
    \label{tab:res_emb1}

\end{table*}

Having established the performance differences between model architectures and different informativeness of graphs, we finally turn to the impact of word embeddings on the models' ability to perform Multihop QA. Use of various word embeddings has a long tradition in QA and other NLP tasks, and can broadly be categorized as \emph{contextualized} or \emph{non-contextualized}. Examples of non-contextualized embeddings include \cite{mikolov2013distributed} and GloVe \cite{pennington2014GloVe}, while contextualized counterparts may be based on LSTM, like ELMO \cite{peters2018deep}, or (more recently) on the transformer model, e.g., BERT \cite{devlin2019bert} or RoBERTa \cite{liu2019roberta}.

While contextualized word embeddings have seen remarkable success in a variety of NLP applications, in large owing to them being able to generate different embeddings for similar surface-level words based on their surroundings, we are still interested in comparing Multihop QA performance given variety of embeddings. In particular, given that a sort of ``contextualization'' is performed for nodes in the constructed entity graphs, it seems reasonable to assume that even non-contextualized embeddings might still yield reasonable performance, or contribute to better QA capabilities when combined with their contextualized counterparts.

We thus conduct a series of experiments where graph nodes are represented using a range of embeddings: GloVe\footnote{300 dimensional GloVe, trained on 840 billion tokens.} (300 dimensions), ELMO\footnote{\url{https://tfhub.dev/google/elmo/3}} (1024 dimensions), RoBERTa\footnote{\url{https://huggingface.co/roberta-large}} (768 dimensions), as well as combinations of GloVe+ELMO, GloVe+RoBERTa, and all three embeddings combined. Table~\ref{tab:res_emb1} shows results for models using graphs of only query and candidate nodes (left numbers), and those using additional reasoning entities (right numbers), respectively.

We observe a clear trend across all models, which consistently perform weakest when using only GloVe, while they exhibit better performance with contextualised embeddings. Noticeably, we observe an edge of ELMO over RoBERTa embeddings when used on their own, and in combinataion with GloVe. While this may seem surprising in light of much current work in NLP, that by and large seems to prefer the (newer) transformer-based embeddings, it nevertheless is in line with recent work on GNNs for QA, which consistently employ ELMO (or a combination of ELMO with other embeddings) as the representation of choice.

Finally, we observer that models generally perform best when employed on graphs that have node representations based on all three embeddings.


\section{Analysis}

\paragraph{Architecture, Graph Structure, Embeddings}

In the previous section, we ran experiments on various RGCN networks for Multihop Question Answering under changing conditions. We varied the underlying model architecture, the provided information in terms of the available graph nodes and relations, and node types of embedding.

One key finding is that our experiments indicate that \emph{information is more important than architecture}. That is to say, we observe that all tested architectures -- EntityGCN, PathGCN, and MashupGCN -- are able to eventually achieve very similar performance in our experiments, at $~68$ answer prediction accuracy. On the other hand, a much larger impact on performance can be attributed to the information that is available to address the question.

Firstly, across architectures, we do see a gain in model performance when using the RGCN over versions omitting this mechanism (Table~\ref{tab:res_arch_rgcn}). This is particularly true for EntityGCN, which gains more than $3$ points of accuracy when making use of its GNN. For PathGCN, this difference is much less pronounced; here, the bi-directional attention layer seems to be able to compensate for much work. However, a small improvement from the GNN can still be observed. It should also be noted that in this particular setting -- ELMO embeddings, for graphs consisting only of query and candidate nodes -- the newly introduced MashupGCN does exhibit a large performance lead over the two original architectures. However, this model does have a larger number of parameters that may account for this difference, which disappears once we start varying the information available to the networks.

On the side of information, we do note a large impact both in terms of graph structure and node embedding type (Tables~\ref{tab:res_graph} and~\ref{tab:res_emb1}).

Firstly, all models seem to be able to gain performance when using the additional nodes and relations introduced by \citet{tangmulti} (row ``+Reason'' in Table~\ref{tab:res_graph}), rather than only question and candidate nodes and relations. The same seems generally true, to some lesser extent, when models have access to additional information based on the background document sentences, rather than to entities only (row ``+sents''). Models are able to achieve their best results when combining reasoning and sentence information in the graph, albeit only when the number of trainable parameters is increased, which renders the comparison unfair to a certain extend\footnote{experiments with larger models in the other settings did not exhibit the same increase in performance.}.

Second, using graphs that encode query, candidate, and reasoning information, we observe an additional large impact of the node embedding type on model performance. Across architectures, performance is worse when using simple GloVe embeddings to represent nodes, and generally increases when using contextualised embeddings. There also seems to be a general benefit of combining both embedding types, as exhibited by columns 4--6 in Table~\ref{tab:res_emb1}, for which both models with and without access to reasoning information exhibit their best respective performances.

\begin{figure}[t]
    \centering
    \includegraphics[width=.90\columnwidth]{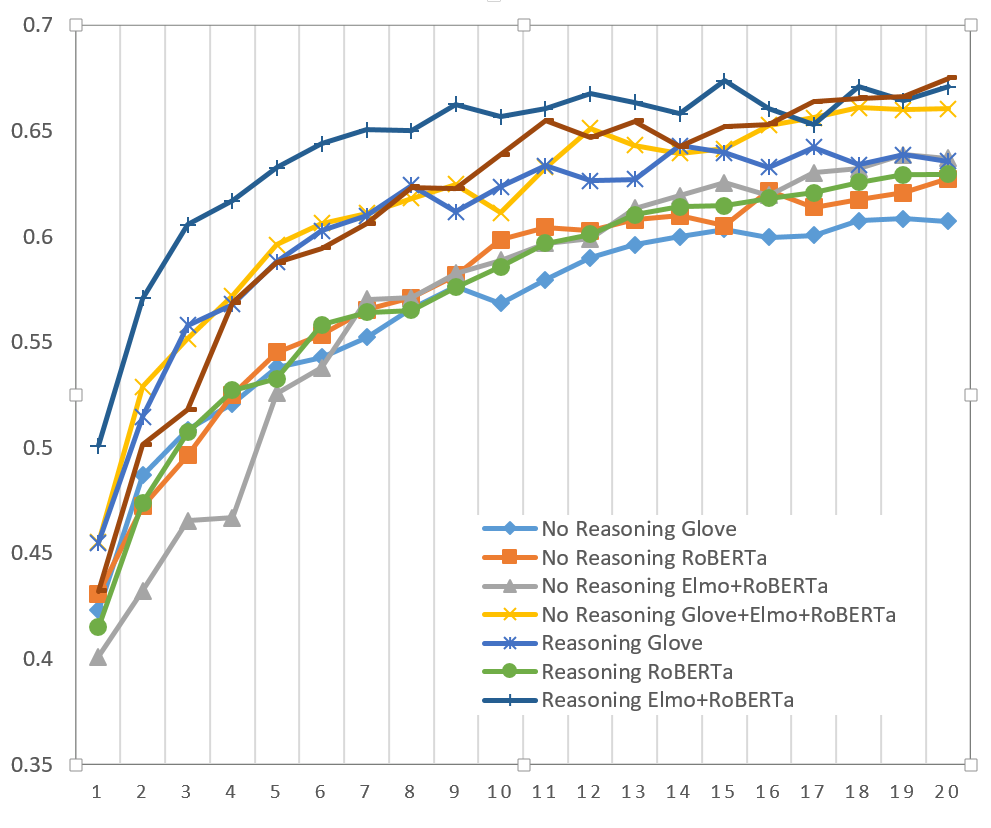}
    \caption{Training curves of tested models with and without reasoning graphs, with various embeddings.}
    \label{fig:training_curve}
\end{figure}

The importance of available information is further illustrated in Figure~\ref{fig:training_curve}. Models with access to reasoning information generally outperform their counterparts even from the very beginning of training, and how the node embedding type has further impact on overall model performance.

We finally note that both EntityGCN and MashupGCN are eventually able to reach answer accuracies of around $68\%$, indicating again that a careful choice of information that is available to the model might be more important than introducing newer, more complicated architectures\footnote{PathGCN seems to under-perform here, especially looking at the original paper's numbers. While the trends and results in terms of impact of information hold, we expected this model to be generally stronger. Unfortunately, code is not publicly available, and authors did not respond to contact.}.

\paragraph{Error Analysis}

We performed a qualitative error analysis on a few selected models to evaluate the types of errors performed by the models. First, a random subset of 50 wrongly classified items from the model with relations from \citet{de2019question} and ELMo embedding (base model) were selected. The types of errors in this model were then compared to the errors in the models with additional relations. In order to do that, we selected the subset of 50 random items which were wrongly classified by the base model but got corrected in the models with additional relations. 

The error analysis shows that the base model makes a wider range of errors, including the following types: (i) overspecification/underspecification; (ii) choosing an entity that is less hops away from the entity mentioned in the query; (iii) wrong entities that are of the same category as the correct answer; (iv) entities in the same topic as the correct answer but not the correct entity category. 

For example, an item is of type (i) if it answers the query ``location christopher street" with \textit{New York} instead of \textit{Greenwich Village}. While technically not false, the answer is underspecified. 

The models with additional sentence and path relations perform in a similar manner to each other with regard to the error types. They both present problems of types (iii) and (iv), but very rarely of the other two types. For example, these models make many mistakes where the answer to the query ``parent\_taxon geastrales" is \textit{phallomycetidae} but the model with sentence relations predicts \textit{rhizoctonia}, which is also a funghi but not the correct type (error type iii). On the other hand, the error type (iv) appears when the model predicts the answer to the query ``subclass\_of hulme f1" to be \textit{formula 1} instead of the correct answer \textit{automobile}. The predicted answer is in the correct topic as this particular model is used for formula 1 racing, however the answer is not the correct category of an entity. Such error types suggest that the models have learned to find the topic or the category of the correct answer through the reasoning paths, however they still have some problems pinpointing the exact correct answer in some cases. 

\section{Conclusions}
In this paper, we have presented a first comparative study Relational Graph Convolutional Networks for Multihop Question Answering, under varying conditions on two modalities: Model architecture, and information available to the models.

We compared three architectures -- EntityGCN, PathGCN, MashupGCN -- and two types of information -- graph structure, node embedding type -- to assess their impact on QA performance on the WikiHop dataset.

Our experiments indicate that while model architecture does have some influence, a much greater source of performance improvement can be found in choosing the right information that is available to solve the task.

Interestingly, in our evaluation, the overall best performance was reached by an earlier established model, EntityGCN, when using information introduced by an extension of the same, PathGCN.

We hope that the results presented here can give new impulses for research in this area to further investigate how we can make better use of information that is already available, to improve the reasoning capabilities of our models.

\bibliography{acl_bib.bib}
\bibliographystyle{acl_natbib}

\end{document}